\documentclass[sigconf]{acmart}
\AtBeginDocument{%
  }

\setcopyright{acmlicensed}
\copyrightyear{2018}
\acmYear{2018}
\acmDOI{XXXXXXX.XXXXXXX}
\acmConference[Conference acronym 'XX]{Make sure to enter the correct
  conference title from your rights confirmation email}{June 03--05,
  2018}{Woodstock, NY}
\acmISBN{978-1-4503-XXXX-X/2018/06}

\usepackage{graphicx}
\usepackage{booktabs}
\usepackage{amsmath}
\usepackage{colortbl}
\usepackage{multirow}
\usepackage{epigraph}
\usepackage{listings}
\usepackage[most]{tcolorbox}

\begin{document}

\title{GeoExplain: Multimodal Reasoning based on Hierarchy of Visual Information in Street View}

\author{Fenghua Cheng}
\email{fenghua.cheng@uq.edu.au}
\affiliation{%
  \institution{The University of Queensland}
  \city{Brisbane}
  \state{Queensland}
  \country{Australia}
}

\author{Jinxiang Wang}
\email{jinxiang.wang@uq.edu.au}
\affiliation{%
  \institution{The University of Queensland}
  \city{Brisbane}
  \state{Queensland}
  \country{Australia}
}

\author{Sen Wang}
\email{sen.wang@uq.edu.au}
\affiliation{%
  \institution{The University of Queensland}
  \city{Brisbane}
  \state{Queensland}
  \country{Australia}
}

\author{Zi Huang}
\email{helen.huang@uq.edu.au}
\affiliation{%
  \institution{The University of Queensland}
  \city{Brisbane}
  \state{Queensland}
  \country{Australia}
}

\author{Xue Li}
\email{xueli@eecs.uq.edu.au}
\affiliation{%
  \institution{The University of Queensland}
  \city{Brisbane}
  \state{Queensland}
  \country{Australia}
}

\renewcommand{\shortauthors}{Fenghua et al.}
\renewcommand\footnotetextcopyrightpermission[1]{}
\settopmatter{printacmref=false}

\begin{abstract}
Multimodal reasoning is a process of understanding, integrating and inferring information across different data modalities. It has recently attracted surging academic attention. Although there are various tasks for evaluating multimodal reasoning ability, they still have limitations. Reasoning on hierarchical visual clues at different levels of granularity, i.e., local details and global context, is of little discussion, despite its frequent involvement in human reasoning. To bridge the gap, we introduce a challenging dataset, namely GeoExplain, which evaluates explainable geo-localization. Given a street view image, the task is to predict its location and provide a detailed explanation. GeoExplain consists of 40350 panoramas-location-explanation tuples. Each instance contains a set of street-view panoramas, a location on street level, and human-expert explanations describing how the location can be inferred from the visual content of panoramas. Additionally, we present a multimodal and multilevel reasoning method, namely SightSense which can make predictions and generate a comprehensive explanation. Our analysis and experiments demonstrate its outstanding performance in GeoExplain. The dataset can be accessed via: \url{https://anonymous.4open.science/r/GeoExplain-A954}.
\end{abstract}

\ccsdesc[500]{Computing methodologies~Natural language generation}

\keywords{Multimodal Reasoning, Multimodal Dataset, Image Geo-Localization, Hierarchical Visual Information}
\begin{teaserfigure}
 \includegraphics[width=\textwidth]{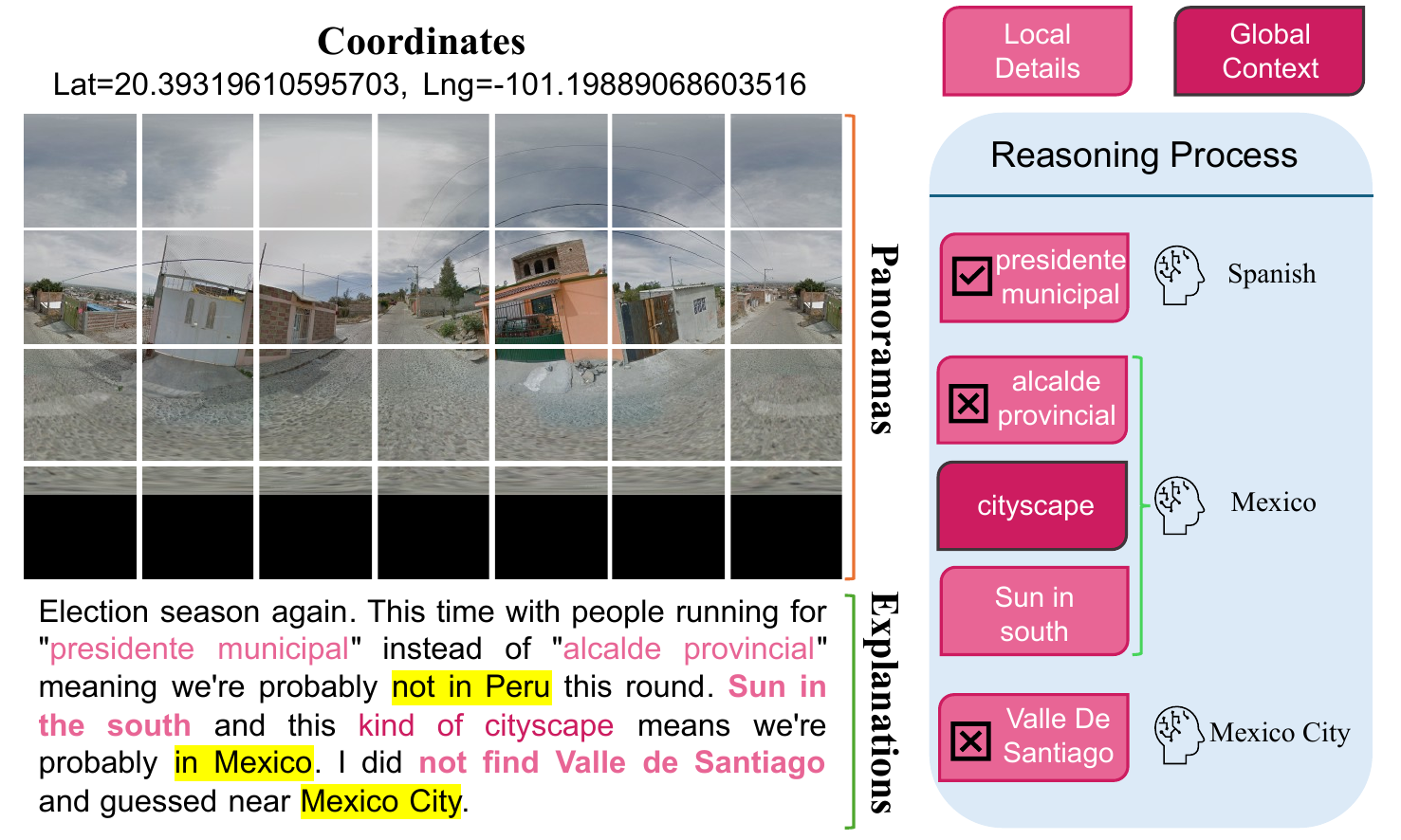}
  \caption{A GeoExplain example: each sample consists of a set of panoramas, location and explanation on how to infer the location. The explanation progressively integrates fine-grained cues (administrative terminology and language) and global context (cityscape and solar direction) to infer the country and further narrow the prediction to a specific city.}
  \Description{A GeoExplain example: panoramas-location-explanation tuple. Each sample consists of a set of panoramas, location and explanation on how to infer the location. Hierarchical visual information, i.e., local details and global context are both necessary in this example.}
  \label{fig:GeoExplain Example}
\end{teaserfigure}


\maketitle

\section{Introduction}
Humans cannot only see the world, but more importantly comprehend the world through it. Observing, associating visual or textual knowledge, making interpretation, and even building a cognitive framework demonstrates a critical and widely used ability of human intelligence, known as multimodal reasoning. Therefore, developing more human-like multimodal reasoning ability is an essential step towards advancing Artificial General Intelligence (AGI) \cite{agi}. 

In recent years, there has been significant enthusiasm for researching on Large Language Model (LLM) \cite{radford2018improving, gpt3, bert, llama} driven by the rise of self-attention mechanism and Transformer architecture \cite{transformer}. We have witnessed the remarkable performance of LLM on Natural Language Understanding \cite{surveyllm1, t5, xlnet}. The great progress prompted the AI community to embark on an exploration of Multimodal Large Language Model (MLLM) \cite{clip, flamingo, blip2} and multimodal reasoning \cite{vqa, multimodalreasoning1} that incorporate visual context as additional input. Despite the success achieved by techniques such as the Chain-of-Thought (CoT) \cite{CoT}, Tree-of-Thoughts (ToT) \cite{ToT}, and Self-Consistency \cite{selfconsistency}, some studies highlighted that state-of-the-art MLLMs still struggle in multimodal reasoning \cite{blink, puzzlegamellm}. 

In response, a diverse range of multimodal reasoning testbeds have been developed, such as Science Question Answering (ScienceQA) \cite{scienceqa}, Geometric Question Answering \cite{geoqa} and GQA \cite{gqa}. Although these testbeds vary in domains and disciplines, we believe they still face the following limitations: 
\begin{itemize}
    \item \textbf{Superficial Reasoning Process}: These testbeds were relatively simple, requiring only superficial reasoning process and limited external knowledge which cannot embody the diverse, hierarchical and complex nature of human reasoning. This limitation weakens the practical grounding of expert-level multimodal reasoning. 
    \item \textbf{Insufficient Visual Context at Different Granularity}: Existing multimodal reasoning testbeds primarily focus on reasoning over global visual context, often neglecting local visual clues. Identifying tiny visual clues and extracting meaningful information from them is also an intuitive aspect of human cognition. Lack of multi-level visual understanding restricts reasoning performance on detail-oriented tasks. 
\end{itemize}

To this end, we introduce a new multimodal dataset for explainable geo-localization: GeoExplain. Unlike existing exploration of image geo-localization \cite{imagegeolocalization, img2loc} that only requires predicting coordinates of the location in a black box manner, GeoExplain emphasizes explanations that reveals how the location can be inferred from the visual evidence as illustrated in Figure \ref{fig:GeoExplain Example}. Participants must analyze hierarchical visual elements at different levels of granularity, ranging from global context like landscape and local details like car plate. The complexity reflects the multifaceted nature of real-world reasoning, making it a challenging endeavor even for humans. Despite recent progress in multimodal reasoning, explainable street-view geo-localization remains insufficiently explored. 

GeoExplain is collected from the image geo-localization game GeoGuessr \cite{GeoGuessr}. Each sample in GeoExplain includes a set of panoramic images, the location on street level and an explanation of how the exact location can be inferred from the image contents. We remove locations that are easy to recognize to make GeoExplain an expert-level task even for humans. We also propose a three-stage method, SightSense, specially designed for hierarchical multimodal reasoning to leverage local and global visual understanding and external knowledge. More specifically, in the first stage, the model identifies fine-grained visual clues via an open-vocabulary object detection. The detected objects are then used to retrieve knowledge snippets in the second stage. Finally, these elements and global context are fused through an adapter, enabling the model to generate a comprehensive response. Our experiments demonstrate the effectiveness of SightSense on GeoExplain, outperforming other state-of-the-art baselines. 

We summarize this work's contribution as follows: 

\begin{itemize}
    \item We propose a novel multimodal reasoning dataset for explainable geo-localization: GeoExplain. GeoExplain provides human-expert natural-language explanation of how a location can be inferred from the visual contents.
    \item We propose a new three-stage method SightSense for hierarchical multimodal reasoning to identify visual hints from different levels of granularity, retrieve and leverage external knowledge. 
    \item We also conduct experiments on GeoExplain. The results demonstrate that our approach surpasses other baselines. 
\end{itemize}

\section{Related Works}

\subsection{Image Geo-Localization}

Recent years have witnessed remarkable advancements in the field of image geo-localization, a subdomain of computer vision. The task of image geo-localization involves estimating the precise geographic coordinates where the photograph was taken. Recent developments in image geo-localization can be categorized into three main approaches: retrieval-based method, regression-based method and LLM-based method. 

The retrieval-based method \cite{retrieval1, retrieval2, retrieval3, retrieval4, retrieval5} searches for the most similar photo within a large-scale photo set in which each photo is annotated with its exact geographical coordinates. The coordinates of the best-matched image are then identified as the result. The regression-based method \cite{streetclip, geoclip, classification2, classification3, classification4, classification5} approaches the geo-localization task as an earth segments regression problem, employing deep learning networks in a black-box manner. Recent LLM-based method \cite{llmgeolocalization1, llmgeolocalization2, llmgeolocalization3, llmgeolocalization4} utilizes the reasoning ability obtained from large-scale training corpus to generate prediction in natural language. 

Despite strong progress, each approach still faces notable challenges: matching images from different viewpoints for retrieval-based method, lack of interpretability for regression-based method and hallucination for LLM-based method \cite{llmhallucination}. 


\subsection{Visual Language Modelling}

Scaling up the size of models' parameters and training data has significantly enhanced the capabilities of attention-based Large Language Models (LLMs) \cite{radford2018improving, gpt3, bert}, enabling them to surpass human performance in various generative tasks in Natural Language Processing (NLP), such as question answering \cite{llmonqa1, llmonqa2} and text summarization \cite{llmontenxsummarization1, llmontenxsummarization2}. However, most of them were limited by inability to read visual information. Conversely, Large Vision Models (LVMs) \cite{lvm1, lvm2, lvm3} can see clearly but cannot talk. Building on the success of both LLM and LVM, significant research efforts have been dedicated to extending the impressive zero-shot capabilities of LLMs to the visual domain by incorporating visual modalities of LVM into generative tasks, which is called Multimodal Large Language Model (MLLM), such as GPT-4 \cite{gpt4}, LLaVA \cite{llava} and Llama \cite{llama3}. An MLLM can bridge effective connections between different gaps, thus succeeding in various multimodal tasks, including 1) image-grounded question-answering dialogue system \cite{imagegroundedqa1, imagegroundedqa2, imagegroundedqa3}, 2) image-grounded chit-chat dialogue system \cite{imagegroundedchitchat1, imagegroundedchitchat2}, 3) visual-evidence-embedded dialogue system \cite{visualevidenceembedded1, visualevidenceembedded2}, 4) image-response dialogue system \cite{imageresponse1, imageresponse2}, 5) video-grounded question-answering dialogue system \cite{videogroundedqa1, videogroundedqa2, videogroundedqa3}, 6) video-grounded chit-chat dialogue system \cite{videogroundedchitchat1, videogroundedchitchat2, videogroundedchitchat3} and 7) video-grounded real-time comment dialogue \cite{videogroundedrealtime1, videogroundedrealtime2}. 

\subsection{Knowledge-Intensive Visual Reasoning}

Except bridging modality gaps and enabling multi-granularity, reducing hallucination remains an evident challenge in developing convincing MLLMs \cite{llmhallucination}. Most notably issues related to hallucination happen to knowledge-intensive visual language tasks, \textit{e.g.}, knowledge-intensive VQA \cite{knowledgevqa1, knowledgevqa2}, multi-hop VQA \cite{multihopvqa1} and visual reasoning \cite{visionreasoning1, visionreasoning2}. 

Knowledge-intensive VQA requires various external expert knowledge which may not be fully covered by a MLLM's internal parametric knowledge to support accurate answer generation. To mitigate hallucination caused by the absence of expert knowledge, Retrieval Augmented Generation (RAG) \cite{rag} leverage a search engine for external knowledge retrieval, assisting answer generation \cite{ragapplication}. Visual reasoning aims to equip machines with the ability to analyze visual information and make decisions or inferences based on that. Studies \cite{blink, puzzlegamellm} have demonstrated visual reasoning abilities in MLLM still falls behind human performance in a variety of tasks, despite the emergence of Chain-of-Thought (CoT) prompting and related techniques \cite{CoT, ToT}. 

\section{GeoExplain}
\label{sec:GeoExplain}

\subsection{Dataset Overview}

GeoExplain is a multimodal dataset for explainable street-view geo-localization and is designed to emphasize reasoning over hierarchical visual cues, including local details and global contextual information. Each instance in GeoExplain corresponds to a real-world location and is represented as a panorama-location-explanation tuple. An instance in GeoExplain consists of a set of street-view panoramas, a street-level location, and a human-expert explanation describing how the location can be inferred from the visual evidence, represented by Equation \ref{equ:GeoExplain-example}. The panorama $\mathbf{P_{i}}=\{I_{i1}, I_{i2}, \dots, I_{im}\}$ is a series of street view images of the location from different angles, $L_{i}$ is the exact location on street level, and $E_{i}$ denotes the explanation. Figure \ref{fig:GeoExplain Example} demonstrates an example of GeoExplain.

\begin{equation}
    G=\{(\mathbf{P_{i}},L_{i},E_{i})\}_{i=1}^{n}
    \label{equ:GeoExplain-example}
\end{equation}

After filtering, GeoExplain contains 5459 locations, 169939 panorama images, and 40350 explanations. The dataset supports the evaluation of both prediction accuracy and reasoning process. The split of dataset can be accessed via our link with train (4367 locations) and test (1092 locations). 

\subsection{Data Source}

GeoExplain is collected from GeoGuessr\footnote{\href{https://www.geoguessr.com/}{GeoGuessr - Let's explore the world!}} daily challenges and corresponding discussion threads on Reddit\footnote{\href{https://www.reddit.com/}{Reddit - The heart of the internet}} posted by expert players. GeoGuessr is an online geo-localization game which challenges players to guess locations around the world from Google Street View Images\footnote{\href{https://www.google.com.au/intl/en-GB/streetview/}{Explore Street View and add your own 360 images to Google Maps}}. In its daily challenge mode, players are offered five locations sampled from around the world to guess. Thanks to its large player base and active community on Reddit, players who accurately point these locations share their experience in the daily challenge discussion threads on Reddit. These discussions provide detailed reasoning about how locations can be identified from observations at different visual granularity. 

\subsection{Data Collection Filtering}

We collect candidate locations from 1217 historical GeoGuessr daily challenges released between June 2021 and September 2024. GeoExplain is then built on this initial pool of 6085 locations. To improve the quality of GeoExplain, we apply filtering strategies at both location and explanation level. 

At the location level, we aim to remove samples that can be solved without reasoning to maintain the expert-level difficulty of GeoExplain and improve diversity among all locations. To this end, we manually filter the initial pool by following steps:

\begin{itemize}
    \item 1) We remove candidate locations that are close to famous landmarks or contain recognizable landmark objects, e.g., Eiffel Tower in Paris or Opera House in Sydney. 
    \item 2) We discard candidate locations where no reasonable and well-supported explanations of how to find the place are posted in the corresponding discussion threads. 
    \item 3) We filter out high-frequency places and ensure they are distributed across different continents, landscapes and both urban and rural areas.
\end{itemize}

At the explanation level, we prioritize explanations that reflect strong evidence-based reasoning by manually filtering steps:

\begin{itemize}
    \item 1) We remove candidate explanations with wrong guesses. Posters usually include their game scores in the discussion. We remove all explanations with game scores lower than 4000 (with max 5000) \footnote{$d = -\frac{S}{10}\ln\!\left(\frac{s}{5000}\right)$, where \(d\) is the geographic distance, \(s\) is the score, and \(S\) is the scale parameter of the scoring function.} 
    \item 2) We remove non-English explanations to maintain language consistency. 
    \item 3) We remove explanations that rely on non-reasoning factors, such as living nearby or having previously finished the same game. 
\end{itemize}

After filtering these locations and explanations, we collect panoramas from Google Street View under multiple viewing directions. These filtering steps help ensure that GeoExplain is a high-quality and expert-level dataset for explainable street-view geo-localization. 

\subsection{Data Analysis}

GeoExplain contains in total 5459 locations, 169939 panorama images and 40350 explanations on how to find the location based on images. We endeavor to ensure the richness and diversity of vocabulary in these explanations. The locations in GeoExplain are distributed across all continents, covering a wide range of terrains as shown in Table \ref{tab:Location Distribution across All Continents}. They include diverse landscapes such as mountains, deserts, forests, coastlines, and urban environments, ensuring a comprehensive geographical representation. Regarding panorama statistics, each location contains an average of 31.13 street view image snippets, with a maximum of 33 and a minimum of 3. Their sizes vary, ranging from (600 * 280) to (512 * 512). 

\begin{table}[htb]
    \caption{Location Distribution across All Continents}
    \centering
    \begin{tabular}{ccccccc}
    \toprule
        AF & AN & AS & EU & NA & OC & SA \\
        \midrule
        445 & 3 & 1414 & 1680 & 761 & 175 & 981 \\
        \bottomrule
    \end{tabular}
    \label{tab:Location Distribution across All Continents}
\end{table}

One distinctive characteristic of GeoExplain is its emphasis on hierarchical visual reasoning. Solving a sample often requires combining fine-grained local cues with broader contextual signals. To measure this, we conduct a survey between our dataset GeoExplain and other two similar multimodal reasoning datasets: ScienceQA and GeoQA. Specifically, we randomly sampled 100 examples from GeoExplain, ScienceQA and GeoQA respectively. Volunteers are asked to rate them on a scale from 0 to 2 on the following questions. In order to avoid the bias between raters, we anonymize all samples and averaged ratings from three volunteers. Since, even after anonymization, peers could still easily identify which dataset a sample came from, we randomize the order of all samples and invite three non-peers with no prior knowledge of the basic information of these three datasets to rate them. 

\begin{itemize}
    \item The level whether the answer needs reasoning process;
    \item The level whether the answer requires both local visual cues and global context;
    \item The level whether the answer is reasonable and understandable without external knowledge;
    \item The level of difficulty that the answer can be deduced.
\end{itemize}

\begin{figure}[htb]
    \centering
    \includegraphics[width=\linewidth]{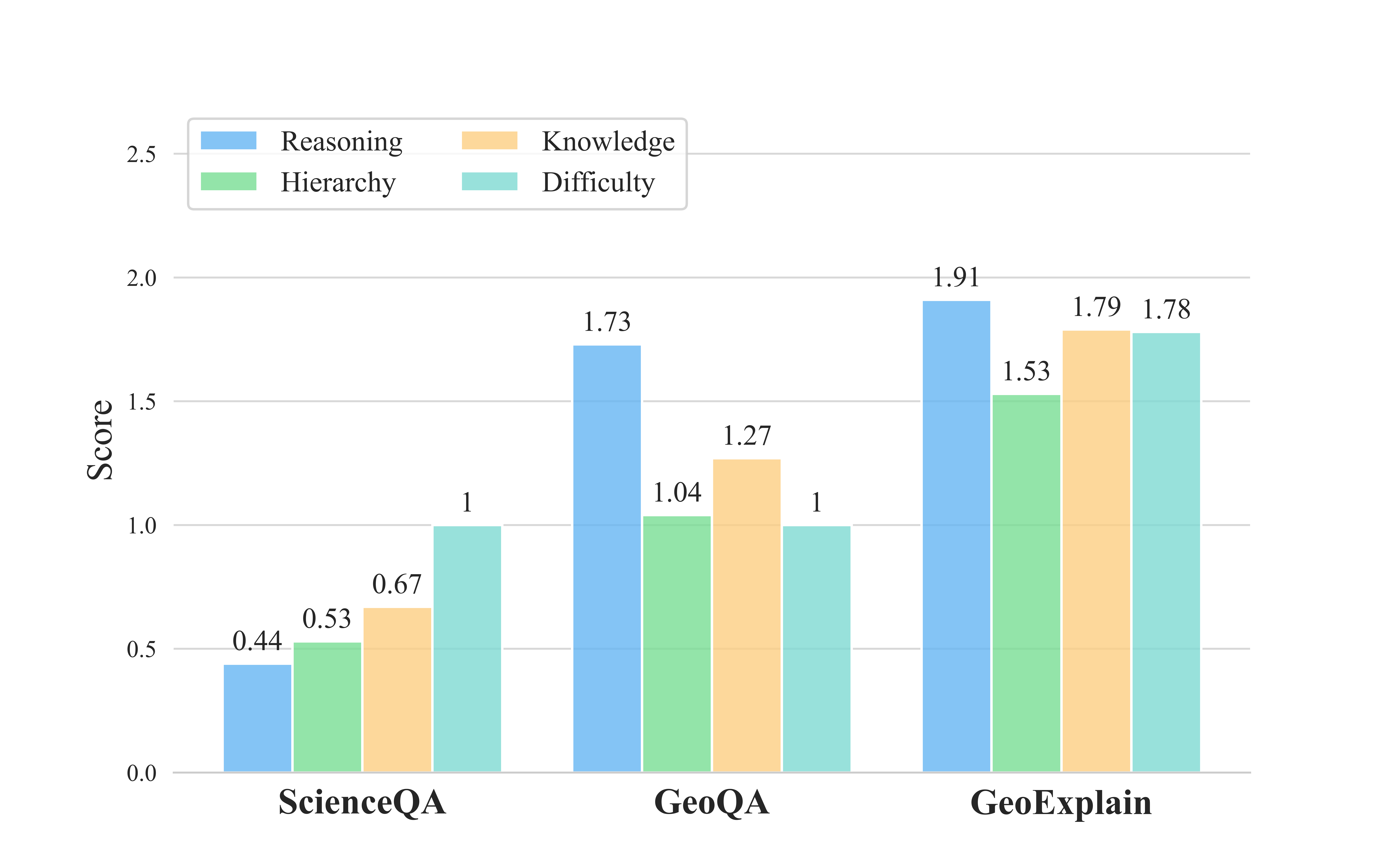}
    \caption{Comparison of Human Analysis across ScienceQA, GeoQA, and GeoExplain. GeoExplain achieves consistently higher scores across all dimensions}
    \label{fig:Human Analysis on GeoExplain and Other Multimodal Reasoning Dataset}
\end{figure}

\begin{figure}
    \centering
    \includegraphics[width=\linewidth]{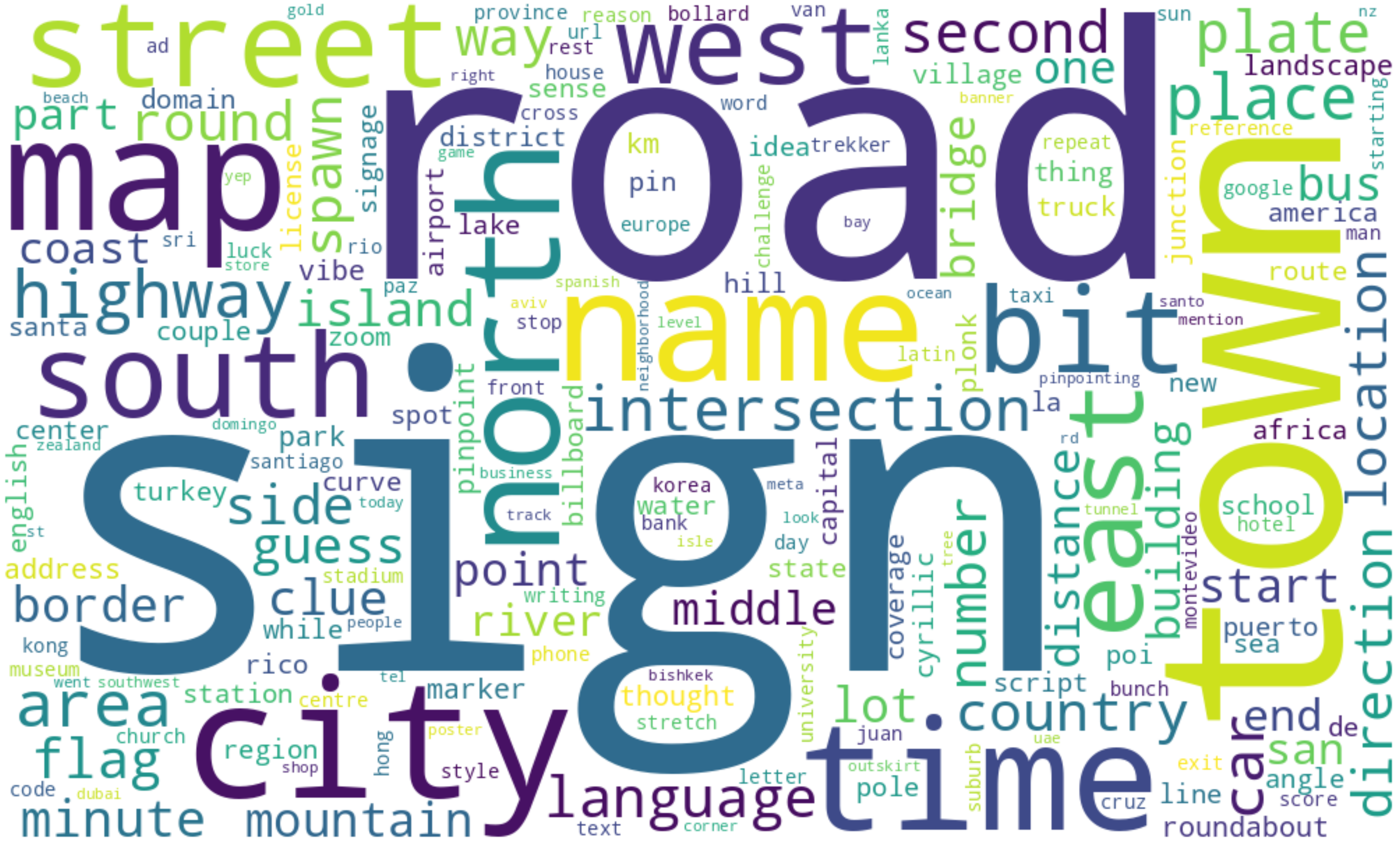}
    \caption{Noun Word Cloud of GeoExplain. Frequently occurring terms such as road, sign, street, city, map, name, and highway indicate that the explanations cover diverse geographic evidence.}
    \label{fig:Noun Word Cloud of GeoExplain}
\end{figure}

From Figure \ref{fig:Human Analysis on GeoExplain and Other Multimodal Reasoning Dataset}, GeoExplain demonstrates the highest level in all four aspects indicating that it demands hierarchical visual reasoning and is at expert-level difficulty. 

We further analyze the explanation in GeoExplain to characterize the reasoning patterns. On average, a location corresponds to 7.39 explanations, with a minimum 1 and maximum 21. The average length of explanations is 203.36 characters. As the Word Cloud Figure \ref{fig:Noun Word Cloud of GeoExplain} shows, the explanations frequently involve cues at different granularity, including sign and city. 

Overall, the analysis indicates that GeoExplain is not only geographically diverse and large in scale, but also challenging and demanding complex reasoning process. 

\subsection{GeoKnowledge}

To incorporate external geographic knowledge, we introduce an auxiliary domain-specific knowledge base, GeoKnowledge. GeoKnowledge consists of image-snippet pairs indexed by country where the snippets tell how to identify a country from visual cues in the image. GeoKnowledge can be represented by Equation \ref{equ:GeoKnowledge-example}, where $KI_{j}$ is an image of country-specific object and $S_{j}$ is the corresponding knowledge snippet.

\begin{equation}
    K=\{(KI_{j}, S_{j})\}_{j=1}^{k}
    \label{equ:GeoKnowledge-example}
\end{equation}

GeoKnowledge is collected from a geographical knowledge website\footnote{\href{https://www.plonkit.net/}{Plonk It}} which is established for solving the image-geolocalization. 

\section{SightSense}

\begin{figure*}[htb]
    \centering
    \includegraphics[width=\textwidth]{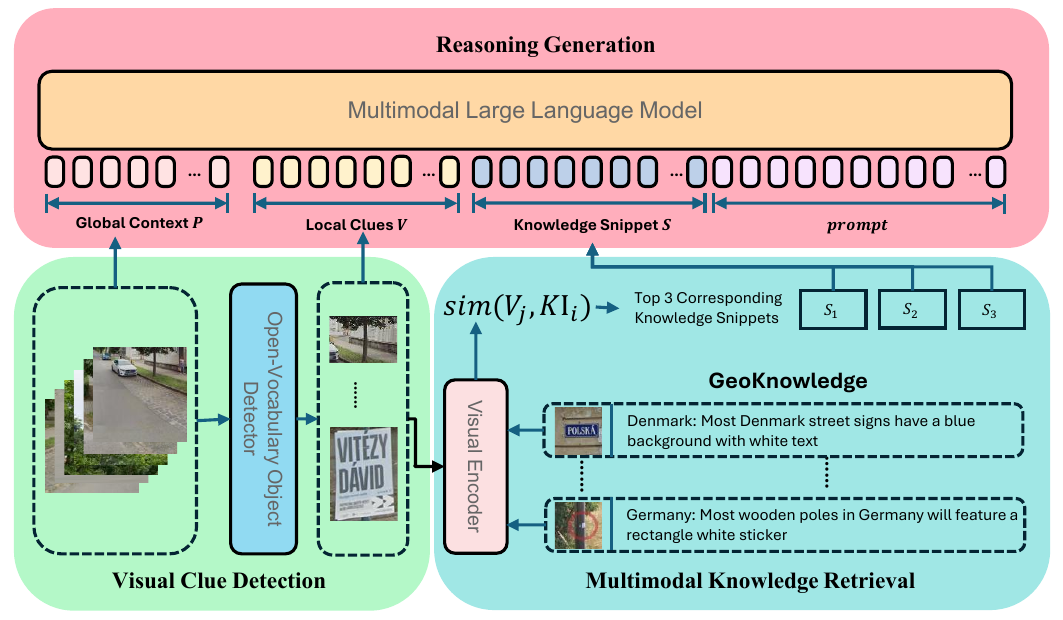}
    \caption{Overview of the SightSense framework. Given a street-view panorama, SightSense first detects fine-grained local visual clues using an open-vocabulary object detector. These clues are then matched against GeoKnowledge through multimodal similarity retrieval to obtain the most relevant geographic knowledge snippets. Finally, the multimodal large language model integrates the global visual context, detected local clues, retrieved knowledge, and task prompt to generate the final geolocation prediction and its corresponding explanation.}
    \label{fig:Architecture of SightSense}
\end{figure*}

To tackle GeoExplain, We propose SightSense, a novel multimodal reasoning approach that integrates hierarchical visual information across multiple levels of granularity including local clues and global context. More specifically, SightSense captures fine-grained visual clues, perceives the coarse-grained landscape, and leverages external knowledge to support reasoning. In this section, we elaborate on the design details of SightSense.

\subsection{Problem Statement}

Suppose we have a GeoExplain set $G=\{(\mathbf{P_{i}},L_{i},E_{i})\}_{i=1}^{n}$ and another GeoKnowledge base $K=\{(KI_{j}, S_{j})\}_{j=1}^{k}$. The goal of SightSense is to learn a generative model $P(L,E|P,K;\theta)$ from $D$ where $\theta$ denotes the parameters of the model. Therefore, we can generate location and corresponding explanation according to Equation \ref{equ:problemstatement}. 

\begin{equation}
    L^{*},E^{*}=\arg\max_{L,E}P(L,E|P,K;\theta)
    \label{equ:problemstatement}
\end{equation}

\subsection{Model Architecture}

In order to fully leverage visual information at different levels of granularity, SightSense uses a three-stage method. Figure \ref{fig:Architecture of SightSense} illustrates the overall workflow. The three stages are Visual Clues Detection, Multimodal Knowledge Retrieval and Reasoning Generation respectively. In the first stage, potentially identifiable local elements are extracted from street view images $\mathbf{P}$ by using an open-vocabulary object detector. The second stage retrieves relevant geographical knowledge $KI,S$ based on these visual clues. Finally, the reasoning stage integrates global context, detected visual clues and retrieved external knowledge to generate informative answer and explanation ${L,E}$. 

\subsection{Visual Clue Detection}

To capture and integrate fine-grained visual details into the reasoning process, in the first stage, we explicitly extract all possible identifiable fine-grained elements $\mathbf{V_{i}}=\{V_{1},V_{2},\dots,V_{in}\}$ by a frozen open-vocabulary object detector according to the Equation \ref{equ:ood}. Each element $V_{ij}$ is a cropped image fragment from original panorama $\mathbf{P_{i}}$, containing one or more specific potentially identifiable local elements. 

\begin{align}
    \mathbf{V_{i}}&=\text{OOD}(\mathbf{P_{i}},prompt) \nonumber \\
    &=\cup_{j=1}^{im}\text{OOD}(I_{j},prompt)
    \label{equ:ood}
\end{align}

Based on analysis of high-frequency nouns in the explanation $E$ and knowledge snippets $S$, we design a comprehensive prompt to capture all potentially identifiable objects. We use Grounding DINO\cite{groundingdino} as our open-vocabulary object detector in this stage. 

\subsection{Multimodal Knowledge Retrieval}

Since GeoExplain is an expert-level task even for humans, relying on the inherent commonsense knowledge in MLLMs obtained from large-scale pretraining, or on reasoning techniques like \cite{CoT, ToT}, is insufficient for generating accurate answers. To alleviate potential hallucinations produced by LLM, we have to explicitly incorporate external geographical knowledge into the reasoning process. Accordingly, in the second stage, we aim to retrieve informative knowledge snippets from GeoKnowledge in a multimodal way to assist the reasoning process, inspired by \cite{rag}. 

Based on the visual cues $\mathbf{V_{i}}$ detected in the first stage, we retrieve top 3 similar object images $KI$ in GeoKnowledge by cosine similarity of image embeddings and exploit their knowledge snippets as assisting external knowledge. Meanwhile, corresponding potentially identifiable objects are chosen as assisting local cues to participate in the reasoning process. The retrieval process can be formulated as Equation \ref{equ:knowledge retrieval}.

\begin{align}
    \mathbf{S^{*}}&=\{S_{i_{1}^{*}}, S_{i_{2}^{*}}, S_{i_{3}^{*}}\} \\
    \mathbf{V^{*}}&=\{V_{j_{1}^{*}}, V_{j_{2}^{*}}, V_{j_{3}^{*}}\} \\
    i_{1}^{*}, i_{2}^{*}, i_{3}^{*}, j_{1}^{*}, j_{2}^{*}, j_{3}^{*}&=\arg\operatorname{Top\text{-}3}_{i,j}\frac{\text{VE}(KI_{i})\cdot \text{VE}(V_{j})}{|\text{VE}(KI_{i})||\text{VE}(V_{j})|}
    \label{equ:knowledge retrieval}
\end{align}

In Equation \ref{equ:knowledge retrieval}, $KI$ denotes an image of country-specific object from GeoKnowledge, and $V$ represents detected visual clues obtained in the first stage. To embed the images, we employ a visual encoder, denoted as $\text{VE}$. In our work, we adopt CLIP-ViT \cite{clip} as the visual encoder with an embedding dimension of 768. 

\subsection{Reasoning Generation}

In the reasoning generation stage, we integrate hierarchical visual information and external geographical knowledge. There are two types of input in a sequence. More specifically, for a specific sample, we first concatenate the global context, i.e., the thumbnail image of the panoramas, and the image of assisting local clues to construct the image prompt. Then we use retrieved knowledge snippets to construct the text prompt. 

In text prompt, we explicitly instruct the MLLM to reason based on provided global context, local clues and relevant external knowledge. We design a standard answer format: "PLACE \{COUNTRY, STATE, CITY, STREET\}. EXPLANATION." This ensures that models can make well-structured prediction for evaluation. Even when using un-finetuned baseline models in experiments, the prompt can instruct the model to attempt to generate an explanation. After constructing both image and text prompt, we feed them into a pre-trained MLLM to perform reasoning and generate final answer. 

\subsection{Training Policy}

We fine-tune the MLLM in the reasoning generation stage by adopting a straightforward autoregressive language modeling objective with optimizing Cross-Entropy Loss. The training objective can be formulated as Equation \ref{equ:train loss} where $x_{t}$ denotes the ground truth token at position $t$. 

\begin{equation}
    \text{Loss}=-\sum_{t=1}^{T}logP(x_{t}|x_{1},\dots,x_{t-1})
    \label{equ:train loss}
\end{equation}

\section{Experiments}

We conduct comprehensive experiments comparing SightSense with state-of-the-art baselines on GeoExplain dataset to evaluate their performance. Additionally, we perform an independent evaluation of the knowledge-retrieval module and invite a human expert to predict 100 randomly selected samples from GeoExplain. In this section, we elaborate on our experimental setup and analyze the results through both quantitative metrics and qualitative case studies. 


\subsection{Experimental Setup}

SightSense utilizes Grounding DINO \cite{groundingdino} as the open-vocabulary object detector in the visual clue detection stage, CLIP-ViT \cite{clip} as the visual encoder in the multimodal knowledge retrieval stage and Qwen2VL-7B-Instruct \cite{Qwen2VL} as the base reasoning model in reasoning generation stage. We fine-tune only parameters in Qwen2VL and other baselines, while keeping the parameters of models in Grounding DINO and CLIP-ViT frozen. 


\subsection{Evaluation Metrics}

We refer to the location and explanations generated by SightSense and other baselines as candidate location and explanations. Evaluation of the GeoExplain comprises two aspects: (1) the accuracy of the candidate location, and (2) the textual similarity between model-generated candidate explanations and human-written references which is measured by both automatic metrics and human evaluation. 

\subsubsection{Accuracy of Location}

We evaluate accuracy of location in four hierarchical levels: (COUNTRY, STATE, CITY, STREET). In case that some countries may have different names, we apply fuzzy match\footnote{\href{https://github.com/rapidfuzz/RapidFuzz}{RapidFuzz} and \href{https://github.com/pycountry/pycountry}{pycountry}}. Accuracy is computed as Equation \ref{equ:accuracy}, where $\text{Match}$ is the fuzzy match function, $c_{i}$ and $gt_{i}$ denote the candidate and ground-truth location respectively. Because of large amount of label "unnamed road" in the street-level evaluation, we removed all samples labeled "unnamed road" and for those samples, accuracy was computed only at higher levels to avoid the effect of long-tail distribution.

\begin{equation}
    \text{Acc}=\frac{\sum_{i=1}^{n}\text{Match}(c_{i},gt_{i})}{n}
    \label{equ:accuracy}
\end{equation}

\subsubsection{Explanation: Automatic Metrics}

To measure the similarity between candidate and reference explanations, we employ two aspects of automatic metrics. 

\begin{itemize}
    \item \textbf{Relevance: }We use (1) BLEU \cite{bleu}, (2) ROUGE-L \cite{rouge}, (3) CIDEr \cite{cider}, (4) Meteor \cite{meteor} and (5) BERTScore \cite{bertscore} to evaluate the relevance between candidate and reference explanations. 
    \item  \textbf{Diversity: }We use (1) Dist-1 and (2) Dist-2 \cite{dist} to measure the diversity of candidate explanations themselves. 
\end{itemize}

\begin{table*}[htb]
    \caption{Automatic evaluation on candidate explanation. Bold number denotes the highest metric value among all baselines and underlying number denotes the highest metric value for each type. $\uparrow$, $\downarrow$ and $-$ indicate how the metric value change for corresponding models when introducing more modules.}
    \centering
    \resizebox{\linewidth}{!}{
    \begin{tabular}{c|c|cccccc|cc}
      \toprule[0.5mm]
      \textbf{Type} & \textbf{Methods} & \textbf{BLEU\_3} & \textbf{BLEU\_4} & \textbf{ROUGE\_L} & \textbf{CIDEr} & \textbf{METEOR} & \textbf{BERTScore} & \textbf{distinct\_1} & \textbf{distinct\_2} \\
      \midrule
      \midrule
      \multirow{6}{*}{\rotatebox{90}{\textit{P}}}
      &LLaVA-1.5-7B & 2.99 & 0.91 & 18.55 & 1.17 & \underline{28.84} & \underline{84.85} & 6.08 & 21.34 \\
      &GPT-4o-mini & 1.53 & 0.00 & 13.95 & 0.22 & 19.44 & 82.59 & 5.32 & 26.30 \\
      &DeepSeek-VL2-tiny & 0.07 & 0.00 & 1.34 & 0.06 & 6.68 & 79.63 & \underline{14.51} & \underline{32.54} \\
      &smileGeo & 0.58 & 0.00 & 7.57 & 0.08 & 10.15 & 79.97 & 12.24 & 28.55 \\
      &GeoReasoner & 1.39 & 0.09 & 9.19 & 0.15 & 17.87 & 81.33 & 7.86 & 25.80 \\
      \cmidrule{2-10}
      &\textbf{\textit{SightSense\_wKVC (ours)}} & \underline{3.71} & \underline{1.92} & \underline{21.13} & \underline{5.51} & 22.71 & 84.46 & 8.57 & 18.45 \\
      \midrule
      \midrule
      \multirow{4}{*}{\rotatebox{90}{\textit{P+VC}}}
      &LLaVA-1.5-7B-VC & 3.00 $\uparrow$ & 0.90 $\downarrow$ & 18.85 $\uparrow$ & 1.03 $\downarrow$ & 28.29 $\downarrow$ & 84.98 $\uparrow$ & 5.02 $\downarrow$ & 17.70 $\downarrow$ \\
      &GPT-4o-mini-VC & 2.12 $\uparrow$ & 0.56 $\uparrow$ & 12.62 $\downarrow$ & 0.33 $\uparrow$ & 23.76 $\uparrow$ & 82.88 $\uparrow$ & 6.92 $\uparrow$ & 28.90 $\uparrow$ \\
      &DeepSeek-VL2-tiny-VC & 0.11 $\uparrow$ & 0.04 $\uparrow$ & 1.75 $\uparrow$ & 0.26 $\uparrow$ & 9.11 $\uparrow$ & 79.79 $\uparrow$ & \textbf{\underline{18.29}} $\uparrow$ & \textbf{\underline{35.32}} $\uparrow$ \\
      \cmidrule{2-10}
      &\textbf{\textit{SightSense\_wK (ours)}} & \underline{6.18} $\uparrow$ & \underline{4.18} $\uparrow$ & \underline{26.88} $\uparrow$ & \underline{6.82} $\uparrow$ & \underline{29.81} $\uparrow$ & \underline{86.18} $\uparrow$ & 9.39 $\uparrow$ & 19.50  $\uparrow$\\
      \midrule
      \midrule
      \multirow{4}{*}{\rotatebox{90}{\textit{P+VC+K}}}
      &LLaVA-1.5-7B-KVC & 5.44 $\uparrow$ & 2.03 $\uparrow$ & 20.00 $\uparrow$ & 1.12 $\uparrow$ & 29.35 $\uparrow$ & 85.39 $\uparrow$ & 4.58 $\downarrow$ & 15.37 $\downarrow$ \\
      &GPT-4o-mini-KVC & 2.45 $\uparrow$ & 0.40 $\downarrow$ & 13.32 $\uparrow$ & 0.37 $\uparrow$ & 22.42 $\downarrow$ & 82.88 $-$ & 7.49 $\uparrow$ & 28.16 $\downarrow$ \\
      &DeepSeek-VL2-tiny-KVC & 1.16 $\uparrow$ & 0.27 $\uparrow$ & 5.46 $\uparrow$ & 0.56 $\uparrow$ & 12.47 $\uparrow$ & 83.36 $\uparrow$ & \underline{13.49} $\downarrow$ & \underline{33.87} $\downarrow$ \\
      \cmidrule{2-10}
      &\textbf{\textit{SightSense (ours)}} & \textbf{\underline{8.93}} $\uparrow$ & \textbf{\underline{5.09}} $\uparrow$ & \textbf{\underline{31.58}} $\uparrow$ & \textbf{\underline{10.43}} $\uparrow$ & \textbf{\underline{36.66}} $\uparrow$ & \textbf{\underline{87.71}} $\uparrow$ & 9.00 $\downarrow$ & 19.23 $\downarrow$ \\
      \bottomrule[0.5mm]
    \end{tabular}}
    \label{tab:Automatic evaluation on candidate explanation}
\end{table*}

\subsubsection{Explanation: Human Evaluation}

Considering the complexity involved in explanation generation, automatic metrics may be insufficient for reliably evaluating reasoning quality and the adequacy of supporting evidence. Therefore, human evaluation is adopted for a more comprehensive analysis. More specifically, we randomly select 100 samples from the test set of GeoExplain. Three voluntary experts in GeoGuessr are asked to rate candidates generated by SightSense and other baselines on a scale from 0 to 2 in following questions. 

\begin{itemize}
    \item \textbf{Fluency: }whether the explanation is fluent and natural. 
    \item \textbf{Evidence: }whether identifiable visual evidence is explicitly mentioned in the explanation. 
    \item \textbf{Knowledge: }whether the explanation incorporates relevant external geographical knowledge or commonsense. 
    \item \textbf{Reasoning: }whether the reasoning process connects the evidence with external knowledge to generate a logically sound, correct, and understandable explanation. 
    \item \textbf{Overall Quality: }overall quality of the generated explanation. 
\end{itemize}

\subsubsection{Knowledge Retrieval Metrics}

For each sample, the knowledge-retrieval module retrieves 3 knowledge snippets, each indexed by a country. By comparing the country labels of the retrieved snippets with the ground truth, we compute common retrieval metrics, including Precision@3, Hit@3 and MRR@3. 

\subsection{Baselines}

We compare SightSense with several state-of-the-art baselines in multimodal reasoning. We validate the effectiveness of hierarchical visual information in the reasoning process even for other baselines by transplanting the first two stages of our pipeline into them. This enables a fair comparison and highlights the generalizability of our approach. Furthermore, we classify the baselines into three different types according to information they receive during generating explanations. Note we do not fine-tune on models with * due to their closed-source nature or inapplicable training policy. 

\noindent \textbullet \; \textbf{Panorama Only: }In this type, we use SightSense\_wKVC which is a simplified version of SightSense, removing the Visual Clue Detection and Knowledge Retrieval. We selected both geo-localization-specific methods and well-established and widely recognized state-of-the-art MLLMs with comparable model sizes: (1) GeoCLIP \cite{geoclip}; (2) StreetCLIP \cite{streetclip}; (3) smileGeo \cite{llmgeolocalization2}; (4) GeoReasoner \cite{llmgeolocalization4}; (5) LLaVA-1.5-7B \cite{llava1.5}; (6) GPT-4o-mini* and (7) DeepSeek-VL2-tiny \cite{deepseekvl2}. These models serve as strong and representative baselines for evaluating the performance of our approach.

\noindent \textbullet \; \textbf{Panorama + Visual Clue: }Similarly, we fine-tune SightSense\_wK which removes only the Knowledge Retrieval from SightSense. Meanwhile, we transplant the Visual Clue Detection stage into other MLLM baselines. We denote them as (1) LLaVA-1.5-7B-VC; (2)GPT-4o-mini-VC* and (3) DeepSeek-VL2-tiny-VC. 

\noindent \textbullet \; \textbf{Panorama + Visual Clue + Knowledge: }We use SightSense with all components enabled. We refer to the baseline models with all three modules as (1) LLaVA-1.5-7B-KVC; (2) GPT-4o-mini-KVC* and (3) DeepSeek-VL2-tiny-KVC. 

\subsection{Main Results}

We summarize results of location accuracy in Table \ref{tab:Accuracy of candidate location}, automatic metrics for explanation in Table \ref{tab:Automatic evaluation on candidate explanation} and human evaluation for explanation in Table \ref{tab:Human Evaluation on candidate explanation}. The standalone assessment of knowledge retrieval module is in Table \ref{tab:Standalone Assessment of Knowledge Retrieval Module}. 

\begin{table}[htb]
    \caption{Accuracy of candidate location in \%. Denotations are the same as in Table \ref{tab:Automatic evaluation on candidate explanation}}
    \centering
    \resizebox{\linewidth}{!}{
    \begin{tabular}{c|c|cccc}
      \toprule[0.5mm]
      \textbf{Type} & \textbf{Methods} & \textbf{Country} & \textbf{State} & \textbf{City} & \textbf{Street} \\
      \midrule
      \midrule
      \multirow{8}{*}{\rotatebox{90}{\textit{P}}}
      &LLaVA-1.5-7B & 16.80 & 0.37 & 0.00 & 0.00 \\
      &GPT-4o-mini & 4.58 & 0.55 & 0.00 & 0.00 \\
      &DeepSeek-VL2-tiny & 9.17 & 1.10 & 0.09 & 0.00 \\
      &GeoCLIP & \underline{55.16} & \underline{16.89} & \underline{3.38} & \underline{1.14} \\
      &StreetCLIP & 48.75 & 7.38 & 2.36 & 0.81 \\
      &smileGeo & 23.44 & 2.27 & 1.52 & 0.14 \\
      &GeoReasoner & 25.81 & 2.46 & 1.13 & 0.54 \\
      \cmidrule{2-6}
      &\textbf{\textit{SightSense\_wKVC (ours)}} & 30.08 & 6.68 & 1.60 & 0.09 \\
      \midrule
      \midrule
      \multirow{4}{*}{\rotatebox{90}{\textit{P+VC}}}
      &LLaVA-1.5-7B-VC & 16.62 $\downarrow$ & 1.38 $\uparrow$ & 0.00 $-$ & 0.00 $-$ \\
      &GPT-4o-mini-VC & 11.38 $\uparrow$ & 0.92 $\uparrow$ & 0.00 $-$ & 0.00 $-$ \\
      &DeepSeek-VL2-tiny-VC & 29.79 $\uparrow$ & \underline{5.32} $\uparrow$ & 0.09 $-$ & 0.00 $-$ \\
      \cmidrule{2-6}
      &\textbf{\textit{SightSense\_wK (ours)}} & \underline{33.46} $\uparrow$ & 5.27 $\downarrow$ & \underline{2.41} $\uparrow$ & \underline{0.51} $\uparrow$ \\
      \midrule
      \midrule
      \multirow{4}{*}{\rotatebox{90}{\textit{P+VC+K}}}
      &LLaVA-1.5-7B-KVC & 25.80 $\uparrow$ & 4.91 $\uparrow$ & 0.10 $\uparrow$ & 0.09 $\uparrow$ \\
      &GPT-4o-mini-KVC & 12.37 $\uparrow$ & 1.01 $\uparrow$ & 0.12 $\uparrow$ & 0.09 $\uparrow$ \\
      &DeepSeek-VL2-tiny-KVC & 32.36 $\uparrow$ & 4.58 $\downarrow$ & 0.37 $\uparrow$ & 0.12 $\uparrow$\\
      \cmidrule{2-6}
      &\textbf{\textit{SightSense (ours)}} & \textbf{\underline{60.97}} $\uparrow$ & \textbf{\underline{23.88}} $\uparrow$ & \textbf{\underline{6.11}} $\uparrow$ & \textbf{\underline{4.55}} $\uparrow$ \\
      \midrule
      \midrule
      \multirow{1}{*}{\rotatebox{90}{}}
      &Human Expert (on 100 Samples) & 75.00 & 42.00 & 15.00 & 8.00 \\
      \bottomrule[0.5mm]
    \end{tabular}}
    \label{tab:Accuracy of candidate location}
\end{table}

\begin{table}[htb]
    \caption{Human Evaluation on candidate explanation. Bold number denotes the highest score among all baselines.}
    \centering
    \resizebox{\linewidth}{!}{
    \begin{tabular}{cccccc}
    \toprule
        \textbf{Method} & \textbf{Fluency} & \textbf{Evidence} & \textbf{Knowledge} & \textbf{Reasoning} & \textbf{Overall} \\
        \midrule
        LLaVA-1.5-7B & 1.01 & 1.05 & 0.77 & 0.61 & 0.67 \\
        GPT4o-mini & \textbf{1.34} & 0.68 & 1.08 & 0.21 & 0.52\\
        DeepSeek-VL2-tiny & 0.89 & 0.03 & 0.01 & 0.03 & 0.1 \\
        smileGeo & 1.05 & 1.13 & 1.07 & 0.89 & 1.10 \\
        GeoReasoner & 0.97 & 0.88 & 1.01 & 0.91 & 1.03 \\
        \textbf{\textit{SightSense (ours)}} & 1.16 & \textbf{1.59} & \textbf{1.67} & \textbf{1.11} & \textbf{1.58}\\
        \bottomrule
    \end{tabular}
    }
    \label{tab:Human Evaluation on candidate explanation}
\end{table}

\begin{table}[htb]
    \caption{Standalone Assessment of Knowledge Retrieval Module in \%.}
    \centering
    
    \begin{tabular}{ccc}
    \toprule
        \textbf{Precision@3} & \textbf{Hit@3} & \textbf{MRR@3} \\
        \midrule
        54.18 & 78.85 & 64.25 \\
        \bottomrule
    \end{tabular}
    
    \label{tab:Standalone Assessment of Knowledge Retrieval Module}
\end{table}

\begin{figure*}
    \centering
    \includegraphics[width=\linewidth]{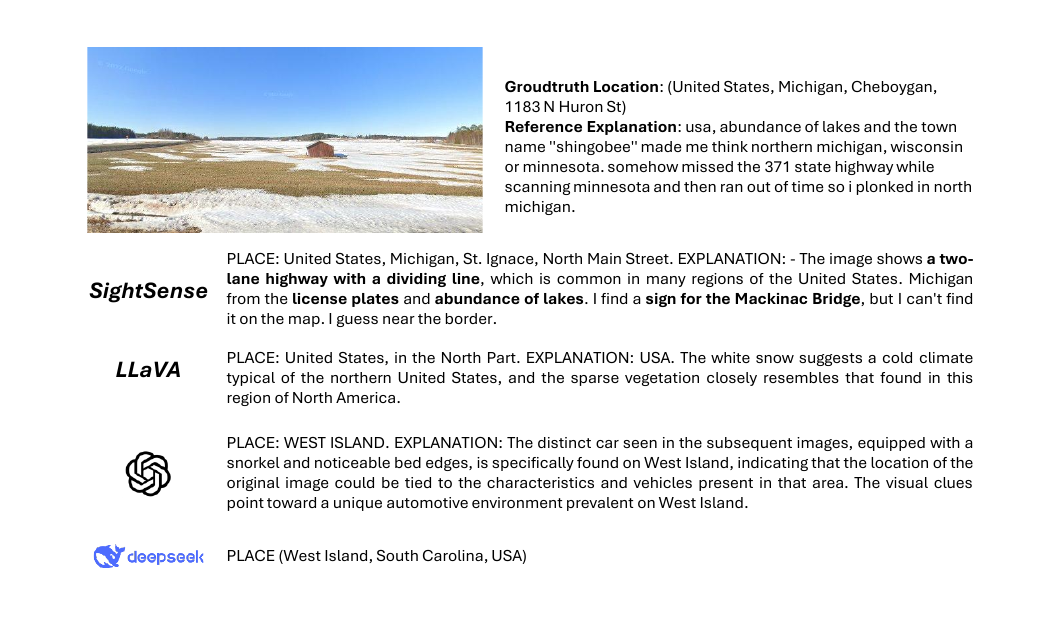}
    \caption{A case study of explanation generated by SightSense and other baselines. SightSense produces a substantially richer informative explanation to narrow down the location to a plausible region in the northern United States. In comparison, the baselines mainly rely on coarse environmental cues or provide less grounded predictions. For visualization, only one downsampled representative view of the panorama is shown, while the models are provided with the complete panoramic input. Therefore, some visual cues mentioned in the generated explanations may not be visible in the displayed image.}
    \label{fig:A case study of explanation generated by SightSense}
\end{figure*}

\noindent \textbullet \; \textbf{Accuracy of Candidate Location: }From Table \ref{tab:Accuracy of candidate location}, SightSense substantially outperforms all other baselines across all four levels, achieving notable gains on the Country and State levels respectively. Meanwhile, within each modality type, the simplified version of SightSense consistently achieves near-best performance except in the Panorama-only type, where SightSense\_wKVC underperforms geo-localization-specified models. However, when all components are enabled, SightSense outperforms even these non-explanation geo-localization-specified methods.This verifies the effectiveness of the architecture and training objective of SightSense and highlights its advantage in addressing the task. The slight accuracy drop is due to misleading of confusing visual clues and potentially wrong retrieved knowledge snippets. We observe some illogical inconsistencies happen, such as accuracy drops at state level while they increase at city level. This suggests that finetuning may cause the model to confuse different administrative tiers. In general, the consistent accuracy gains across all four models further verify the robustness and generalizability of our method. 

On the other hand, almost all baselines struggle at fine-grained levels such as city and street prediction. These numbers still fall behind human expert, indicating GeoExplain is far from solved, particularly at fine-grained level. Moreover, the consistent performance gains from P to P+VC and further to P+VC+K demonstrate that GeoExplain requires hierarchical visual evidence and external geographic knowledge, rather than relying on image matching. This confirms that GeoExplain provides a meaningful testbed for studying reasoning among hierarchical visual evidence in geo-localization. 

\noindent \textbullet \; \textbf{Automatic Metric of Candidate Explanation: }We observe SightSense series still achieves the best performance regarding explanation similarity evaluation across all types in \ref{tab:Automatic evaluation on candidate explanation}. Explanations generated by SightSense show the highest similarity with the reference explanations. The diversity metric is primarily dominated by DeepSeek-VL2 which means SightSense tends to generate a fixed template learned from the GeoExplain dataset. As with location accuracy analysis, we also notice significant increase in similarity metric when introducing Visual Clue Detection and Multimodal Knowledge Retrieval module, proving the effectiveness of our approach. 

\noindent \textbullet \; \textbf{Human Evaluation of Candidate Explanation: }Table \ref{tab:Human Evaluation on candidate explanation} demonstrates the result of human evaluation. Except for Fluency, SightSense achieves the highest score across all other metrics. This indicates that the explanation generated by SightSense provide the strongest evidence, the broadest and the most relevant geographical knowledge, and the most complete reasoning chain. These improvements are achieved by the understanding and integration of hierarchical visual information and external knowledge proposed in SightSense. 

\subsection{Ablation Study}

To better understand the contribution of each stage in SightSense, we provide a brief ablation study. From the comparison across the same model in different types, introducing Visual Clue Detection module results in 8 accuracy increases, 2 decreases and 6 ties. Similarly, introducing Multimodal Knowledge Retrieval module leads to 15 increases and 1 decrease. The comparison demonstrates that integrating hierarchical visual information and external knowledge consistently improves location prediction accuracy. Meanwhile, the consistent gains from P to P+VC and to P+VC+K indicate that GeoExplain requires strong evidence extraction, hierarchical visual reasoning, and geographic knowledge grounding.  

\subsection{Case Study}

To further investigate the performance of SightSense and other baseline models, we present a qualitative case study. As shown in Figure \ref{fig:A case study of explanation generated by SightSense}, SightSense is able to detect the fine-grained visual clues, i.e., Michigan license plate, and perceive the coarse-grained landscape, i.e., abundance of lakes. It then connects such hierarchical visual information with external knowledge to generate the final prediction which is accurate at both country and state level. In contrast, other baselines demonstrate notable limitation. GPT-4o-mini and DeepSeek generate significant hallucination that are inconsistent with visual evidence. LLaVA captures informative clues from the global context but fails to identify detailed visual elements, resulting in a vague location prediction. 

\section{Conclusion}

In this paper, we introduce a multimodal reasoning dataset for explainable street-view geo-localization, GeoExplain, identifying the location of a street view and providing comprehensive explanation. We further propose a three-stage method SightSense which understands hierarchical visual information at multiple levels to solve GeoExplain. Experiments on GeoExplain demonstrate our method outperforms other state-of-the-art baseline models and provide strong generalizability when transplanting to other models. Moreover, no existing model reaches human-level accuracy, which validates the difficulty of GeoExplain and its significance as a dataset toward the development of Artificial General Intelligence. Future work could focus on alleviating long-tail distribution of GeoExplain and reducing negative effects of misleading knowledge snippets. 

\section*{Ethics and Privacy Statement}

GeoExplain is constructed from publicly accessible GeoGuessr challenges, associated public Reddit discussions, and Google Street View imagery. The dataset is intended solely for research on multimodal reasoning and image geo-localization; nevertheless, advances in fine-grained geo-localization may raise privacy and safety concerns if applied to infer sensitive locations or track individuals without consent. We therefore encourage responsible use of the dataset and models, particularly avoiding applications involving personal surveillance, identification, or other privacy-invasive purposes.

\bibliographystyle{ACM-Reference-Format}
\bibliography{sigconf}

@String{Computing = "Computing" }

@String{Computer = "{IEEE} Computer" }

@String{Springer = "Springer-Verlag" }

@article{radford2018improving,
  title={Improving language understanding by generative pre-training},
  author={Radford, Alec and Narasimhan, Karthik and Salimans, Tim and Sutskever, Ilya and others},
  year={2018},
  publisher={San Francisco, CA, USA}
}

@article{gpt3,
  title={Language models are few-shot learners},
  author={Brown, Tom and Mann, Benjamin and Ryder, Nick and Subbiah, Melanie and Kaplan, Jared D and Dhariwal, Prafulla and Neelakantan, Arvind and Shyam, Pranav and Sastry, Girish and Askell, Amanda and others},
  journal={Advances in neural information processing systems},
  volume={33},
  pages={1877--1901},
  year={2020}
}

@inproceedings{bert,
  title={Bert: Pre-training of deep bidirectional transformers for language understanding},
  author={Devlin, Jacob and Chang, Ming-Wei and Lee, Kenton and Toutanova, Kristina},
  booktitle={Proceedings of the 2019 conference of the North American chapter of the association for computational linguistics: human language technologies, volume 1 (long and short papers)},
  pages={4171--4186},
  year={2019}
}

@article{t5,
  title={Exploring the limits of transfer learning with a unified text-to-text transformer},
  author={Raffel, Colin and Shazeer, Noam and Roberts, Adam and Lee, Katherine and Narang, Sharan and Matena, Michael and Zhou, Yanqi and Li, Wei and Liu, Peter J},
  journal={Journal of machine learning research},
  volume={21},
  number={140},
  pages={1--67},
  year={2020}
}

@article{transformer,
  title={Attention is all you need},
  author={Vaswani, Ashish and Shazeer, Noam and Parmar, Niki and Uszkoreit, Jakob and Jones, Llion and Gomez, Aidan N and others},
  journal={Advances in neural information processing systems},
  volume={30},
  year={2017}
}

@article{xlnet,
  title={Xlnet: Generalized autoregressive pretraining for language understanding},
  author={Yang, Zhilin and Dai, Zihang and Yang, Yiming and Carbonell, Jaime and Salakhutdinov, Russ R and Le, Quoc V},
  journal={Advances in neural information processing systems},
  volume={32},
  year={2019}
}

@article{surveyllm1,
  title={Recent advances in natural language processing via large pre-trained language models: A survey},
  author={Min, Bonan and Ross, Hayley and Sulem, Elior and Veyseh, Amir Pouran Ben and Nguyen, Thien Huu and Sainz, Oscar and others},
  journal={ACM Computing Surveys},
  volume={56},
  number={2},
  pages={1--40},
  year={2023},
  publisher={ACM New York, NY}
}

@book{agi,
  title={Artificial general intelligence},
  author={Goertzel, Ben and Pennachin, Cassio},
  volume={2},
  year={2007},
  publisher={Springer}
}

@inproceedings{clip,
  title={Learning transferable visual models from natural language supervision},
  author={Radford, Alec and Kim, Jong Wook and Hallacy, Chris and Ramesh, Aditya and Goh, Gabriel and Agarwal, Sandhini and others},
  booktitle={International conference on machine learning},
  pages={8748--8763},
  year={2021},
  organization={PmLR}
}

@article{flamingo,
  title={Flamingo: a visual language model for few-shot learning},
  author={Alayrac, Jean-Baptiste and Donahue, Jeff and Luc, Pauline and Miech, Antoine and Barr, Iain and Hasson, Yana and Lenc, Karel and Mensch, Arthur and Millican, Katherine and Reynolds, Malcolm and others},
  journal={Advances in neural information processing systems},
  volume={35},
  pages={23716--23736},
  year={2022}
}

@inproceedings{blip2,
  title={Blip-2: Bootstrapping language-image pre-training with frozen image encoders and large language models},
  author={Li, Junnan and Li, Dongxu and Savarese, Silvio and Hoi, Steven},
  booktitle={International conference on machine learning},
  pages={19730--19742},
  year={2023},
  organization={PMLR}
}

@inproceedings{vqa,
  title={Vqa: Visual question answering},
  author={Antol, Stanislaw and Agrawal, Aishwarya and Lu, Jiasen and Mitchell, Margaret and Batra, Dhruv and Zitnick, C Lawrence and Parikh, Devi},
  booktitle={Proceedings of the IEEE international conference on computer vision},
  pages={2425--2433},
  year={2015}
}

@article{multimodalreasoning1,
  title={Enhancing human-like multimodal reasoning: a new challenging dataset and comprehensive framework},
  author={Wei, Jingxuan and Tan, Cheng and Gao, Zhangyang and Sun, Linzhuang and Li, Siyuan and Yu, Bihui and others},
  journal={Neural Computing and Applications},
  volume={36},
  number={33},
  pages={20849--20861},
  year={2024},
  publisher={Springer}
}

@article{CoT,
  title={Chain-of-thought prompting elicits reasoning in large language models},
  author={Wei, Jason and Wang, Xuezhi and Schuurmans, Dale and Bosma, Maarten and Xia, Fei and Chi, Ed and others},
  journal={Advances in neural information processing systems},
  volume={35},
  pages={24824--24837},
  year={2022}
}

@article{ToT,
  title={Tree of thoughts: Deliberate problem solving with large language models},
  author={Yao, Shunyu and Yu, Dian and Zhao, Jeffrey and Shafran, Izhak and Griffiths, Tom and Cao, Yuan and Narasimhan, Karthik},
  journal={Advances in neural information processing systems},
  volume={36},
  pages={11809--11822},
  year={2023}
}

@article{selfconsistency,
  title={Self-consistency improves chain of thought reasoning in language models},
  author={Wang, Xuezhi and Wei, Jason and Schuurmans, Dale and Le, Quoc and Chi, Ed and Narang, Sharan and others},
  journal={arXiv preprint arXiv:2203.11171},
  year={2022}
}

@inproceedings{blink,
  title={Blink: Multimodal large language models can see but not perceive},
  author={Fu, Xingyu and Hu, Yushi and Li, Bangzheng and Feng, Yu and Wang, Haoyu and Lin, Xudong and others},
  booktitle={European Conference on Computer Vision},
  pages={148--166},
  year={2024},
  organization={Springer}
}

@article{puzzlegamellm,
  title={Connecting the dots: Evaluating abstract reasoning capabilities of LLMs using the New York Times Connections word game},
  author={Samadarshi, Prisha and Mustafa, Mariam and Kulkarni, Anushka and Rothkopf, Raven and Chakrabarty, Tuhin and Muresan, Smaranda},
  journal={arXiv preprint arXiv:2406.11012},
  year={2024}
}

@article{scienceqa,
  title={Learn to explain: Multimodal reasoning via thought chains for science question answering},
  author={Lu, Pan and Mishra, Swaroop and Xia, Tanglin and Qiu, Liang and Chang, Kai-Wei and Zhu, Song-Chun and others},
  journal={Advances in Neural Information Processing Systems},
  volume={35},
  pages={2507--2521},
  year={2022}
}

@inproceedings{geoqa,
  title={GeoQA: A Geometric Question Answering Benchmark Towards Multimodal Numerical Reasoning},
  author={Chen, Jiaqi and Tang, Jianheng and Qin, Jinghui and Liang, Xiaodan and Liu, Lingbo and Xing, Eric and Lin, Liang},
  booktitle={Findings of the Association for Computational Linguistics: ACL-IJCNLP 2021},
  pages={513--523},
  year={2021}
}

@inproceedings{gqa,
  title={Gqa: A new dataset for real-world visual reasoning and compositional question answering},
  author={Hudson, Drew A and Manning, Christopher D},
  booktitle={Proceedings of the IEEE/CVF conference on computer vision and pattern recognition},
  pages={6700--6709},
  year={2019}
}

@inproceedings{imagegeolocalization,
  title={Im2gps: estimating geographic information from a single image},
  author={Hays, James and Efros, Alexei A},
  booktitle={2008 ieee conference on computer vision and pattern recognition},
  pages={1--8},
  year={2008},
  organization={IEEE}
}

@inproceedings{img2loc,
  title={Img2Loc: Revisiting image geolocalization using multi-modality foundation models and image-based retrieval-augmented generation},
  author={Zhou, Zhongliang and Zhang, Jielu and Guan, Zihan and Hu, Mengxuan and Lao, Ni and Mu, Lan and Li, Sheng and Mai, Gengchen},
  booktitle={Proceedings of the 47th International ACM SIGIR Conference on Research and Development in Information Retrieval},
  pages={2749--2754},
  year={2024}
}

@inproceedings{retrieval1,
  title={Bridging the domain gap for ground-to-aerial image matching},
  author={Regmi, Krishna and Shah, Mubarak},
  booktitle={Proceedings of the IEEE/CVF International Conference on Computer Vision},
  pages={470--479},
  year={2019}
}

@article{retrieval2,
  title={Spatial-aware feature aggregation for image based cross-view geo-localization},
  author={Shi, Yujiao and Liu, Liu and Yu, Xin and Li, Hongdong},
  journal={Advances in Neural Information Processing Systems},
  volume={32},
  year={2019}
}

@inproceedings{retrieval3,
  title={Where am i looking at? joint location and orientation estimation by cross-view matching},
  author={Shi, Yujiao and Yu, Xin and Campbell, Dylan and Li, Hongdong},
  booktitle={Proceedings of the IEEE/CVF Conference on Computer Vision and Pattern Recognition},
  pages={4064--4072},
  year={2020}
}

@inproceedings{retrieval4,
  title={Coming down to earth: Satellite-to-street view synthesis for geo-localization},
  author={Toker, Aysim and Zhou, Qunjie and Maximov, Maxim and Leal-Taix{\'e}, Laura},
  booktitle={Proceedings of the IEEE/CVF Conference on Computer Vision and Pattern Recognition},
  pages={6488--6497},
  year={2021}
}

@inproceedings{retrieval5,
  title={Vigor: Cross-view image geo-localization beyond one-to-one retrieval},
  author={Zhu, Sijie and Yang, Taojiannan and Chen, Chen},
  booktitle={Proceedings of the IEEE/CVF Conference on Computer Vision and Pattern Recognition},
  pages={3640--3649},
  year={2021}
}

@inproceedings{classification2,
  title={Where in the world is this image? transformer-based geo-localization in the wild},
  author={Pramanick, Shraman and Nowara, Ewa M and Gleason, Joshua and Castillo, Carlos D and Chellappa, Rama},
  booktitle={European Conference on Computer Vision},
  pages={196--215},
  year={2022},
  organization={Springer}
}

@inproceedings{classification3,
  title={Cplanet: Enhancing image geolocalization by combinatorial partitioning of maps},
  author={Seo, Paul Hongsuck and Weyand, Tobias and Sim, Jack and Han, Bohyung},
  booktitle={Proceedings of the European Conference on Computer Vision (ECCV)},
  pages={536--551},
  year={2018}
}

@inproceedings{classification4,
  title={Revisiting im2gps in the deep learning era},
  author={Vo, Nam and Jacobs, Nathan and Hays, James},
  booktitle={Proceedings of the IEEE international conference on computer vision},
  pages={2621--2630},
  year={2017}
}

@inproceedings{classification5,
  title={Planet-photo geolocation with convolutional neural networks},
  author={Weyand, Tobias and Kostrikov, Ilya and Philbin, James},
  booktitle={Computer Vision--ECCV 2016: 14th European Conference, Amsterdam, The Netherlands, October 11-14, 2016, Proceedings, Part VIII 14},
  pages={37--55},
  year={2016},
  organization={Springer}
}

@article{llmgeolocalization1,
  title={Image-Based Geolocation Using Large Vision-Language Models},
  author={Liu, Yi and Ding, Junchen and Deng, Gelei and Li, Yuekang and Zhang, Tianwei and Sun, Weisong and Zheng, Yaowen and Ge, Jingquan and Liu, Yang},
  journal={arXiv preprint arXiv:2408.09474},
  year={2024}
}

@article{llmgeolocalization2,
  title={Swarm Intelligence in Geo-Localization: A Multi-Agent Large Vision-Language Model Collaborative Framework},
  author={Han, Xiao and Zhu, Chen and Zhao, Xiangyu and Zhu, Hengshu},
  journal={arXiv preprint arXiv:2408.11312},
  year={2024}
}

@inproceedings{llmgeolocalization3,
  title={Georeasoner: Reasoning on geospatially grounded context for natural language understanding},
  author={Yan, Yibo and Lee, Joey},
  booktitle={Proceedings of the 33rd ACM International Conference on Information and Knowledge Management},
  pages={4163--4167},
  year={2024}
}

@inproceedings{llmgeolocalization4,
  title={Georeasoner: Geo-localization with reasoning in street views using a large vision-language model},
  author={Li, Ling and Ye, Yu and Jiang, Bingchuan and Zeng, Wei},
  booktitle={Forty-first International Conference on Machine Learning},
  year={2024}
}

@inproceedings{llmonqa1,
  title={Predicting Question-Answering Performance of Large Language Models through Semantic Consistency},
  author={Rabinovich, Ella and Ackerman, Samuel and Raz, Orna and Farchi, Eitan and Tavor, Ateret Anaby},
  booktitle={Proceedings of the Third Workshop on Natural Language Generation, Evaluation, and Metrics (GEM)},
  pages={138--154},
  year={2023}
}

@article{llmonqa2,
  title={Pali: A jointly-scaled multilingual language-image model},
  author={Chen, Xi and Wang, Xiao and Changpinyo, Soravit and Piergiovanni, AJ and Padlewski, Piotr and Salz, Daniel and Goodman, Sebastian and Grycner, Adam and Mustafa, Basil and Beyer, Lucas and others},
  journal={arXiv preprint arXiv:2209.06794},
  year={2022}
}

@inproceedings{llmontenxsummarization1,
  title={On Learning to Summarize with Large Language Models as References},
  author={Liu, Yixin and Shi, Kejian and He, Katherine and Ye, Longtian and Fabbri, Alexander Richard and Liu, Pengfei and Radev, Dragomir and Cohan, Arman},
  booktitle={Proceedings of the 2024 Conference of the North American Chapter of the Association for Computational Linguistics: Human Language Technologies (Volume 1: Long Papers)},
  pages={8639--8656},
  year={2024}
}

@inproceedings{llmontenxsummarization2,
  title={Pegasus: Pre-training with extracted gap-sentences for abstractive summarization},
  author={Zhang, Jingqing and Zhao, Yao and Saleh, Mohammad and Liu, Peter},
  booktitle={International conference on machine learning},
  pages={11328--11339},
  year={2020},
  organization={PMLR}
}

@article{lvm1,
  title={An image is worth 16x16 words: Transformers for image recognition at scale},
  author={Dosovitskiy, Alexey and Beyer, Lucas and Kolesnikov, Alexander and Weissenborn, Dirk and Zhai, Xiaohua and Unterthiner, Thomas and Dehghani, Mostafa and Minderer, Matthias and Heigold, Georg and Gelly, Sylvain and others},
  journal={arXiv preprint arXiv:2010.11929},
  year={2020}
}

@inproceedings{lvm2,
  title={Swin transformer: Hierarchical vision transformer using shifted windows},
  author={Liu, Ze and Lin, Yutong and Cao, Yue and Hu, Han and Wei, Yixuan and Zhang, Zheng and Lin, Stephen and Guo, Baining},
  booktitle={Proceedings of the IEEE/CVF international conference on computer vision},
  pages={10012--10022},
  year={2021}
}

@inproceedings{lvm3,
  title={Cmt: Convolutional neural networks meet vision transformers},
  author={Guo, Jianyuan and Han, Kai and Wu, Han and Tang, Yehui and Chen, Xinghao and Wang, Yunhe and Xu, Chang},
  booktitle={Proceedings of the IEEE/CVF conference on computer vision and pattern recognition},
  pages={12175--12185},
  year={2022}
}

@article{gpt4,
  title={Gpt-4 technical report},
  author={Achiam, Josh and Adler, Steven and Agarwal, Sandhini and Ahmad, Lama and Akkaya, Ilge and Aleman, Florencia Leoni and Almeida, Diogo and Altenschmidt, Janko and Altman, Sam and Anadkat, Shyamal and others},
  journal={arXiv preprint arXiv:2303.08774},
  year={2023}
}

@article{llava,
  title={Visual instruction tuning},
  author={Liu, Haotian and Li, Chunyuan and Wu, Qingyang and Lee, Yong Jae},
  journal={Advances in neural information processing systems},
  volume={36},
  pages={34892--34916},
  year={2023}
}

@article{llama3,
  title={The llama 3 herd of models},
  author={Grattafiori, Aaron and Dubey, Abhimanyu and Jauhri, Abhinav and Pandey, Abhinav and Kadian, Abhishek and Al-Dahle, Ahmad and Letman, Aiesha and Mathur, Akhil and Schelten, Alan and Vaughan, Alex and others},
  journal={arXiv preprint arXiv:2407.21783},
  year={2024}
}

@article{llama,
  title={Llama: Open and efficient foundation language models},
  author={Touvron, Hugo and Lavril, Thibaut and Izacard, Gautier and Martinet, Xavier and Lachaux, Marie-Anne and Lacroix, Timoth{\'e}e and Rozi{\`e}re, Baptiste and Goyal, Naman and Hambro, Eric and Azhar, Faisal and others},
  journal={arXiv preprint arXiv:2302.13971},
  year={2023}
}

@inproceedings{imagegroundedqa1,
  title={Visual dialog},
  author={Das, Abhishek and Kottur, Satwik and Gupta, Khushi and Singh, Avi and Yadav, Deshraj and Moura, Jos{\'e} MF and Parikh, Devi and Batra, Dhruv},
  booktitle={Proceedings of the IEEE conference on computer vision and pattern recognition},
  pages={326--335},
  year={2017}
}

@inproceedings{imagegroundedqa2,
  title={Large-scale pretraining for visual dialog: A simple state-of-the-art baseline},
  author={Murahari, Vishvak and Batra, Dhruv and Parikh, Devi and Das, Abhishek},
  booktitle={European Conference on Computer Vision},
  pages={336--352},
  year={2020},
  organization={Springer}
}

@article{imagegroundedqa3,
  title={Efficient Attention Mechanism for Visual Dialog that can Handle All the Interactions between Multiple Inputs},
  author={Nguyen, Van-Quang and Suganuma, Masanori and Okatani, Takayuki},
  journal={arXiv preprint arXiv:1911.11390},
  year={2019}
}

@inproceedings{imagegroundedchitchat1,
  title={Image-Chat: Engaging Grounded Conversations},
  author={Shuster, Kurt and Humeau, Samuel and Bordes, Antoine and Weston, Jason},
  booktitle={Proceedings of the 58th Annual Meeting of the Association for Computational Linguistics},
  pages={2414--2429},
  year={2020}
}

@inproceedings{imagegroundedchitchat2,
  title={Multi-Modal Open-Domain Dialogue},
  author={Shuster, Kurt and Smith, Eric Michael and Ju, Da and Weston, Jason},
  booktitle={Proceedings of the 2021 Conference on Empirical Methods in Natural Language Processing},
  pages={4863--4883},
  year={2021}
}

@inproceedings{visualevidenceembedded1,
  title={Text is not enough: Integrating visual impressions into open-domain dialogue generation},
  author={Shen, Lei and Zhan, Haolan and Shen, Xin and Song, Yonghao and Zhao, Xiaofang},
  booktitle={Proceedings of the 29th ACM International Conference on Multimedia},
  pages={4287--4296},
  year={2021}
}

@article{visualevidenceembedded2,
  title={Maria: A visual experience powered conversational agent},
  author={Liang, Zujie and Hu, Huang and Xu, Can and Tao, Chongyang and Geng, Xiubo and Chen, Yining and Liang, Fan and Jiang, Daxin},
  journal={arXiv preprint arXiv:2105.13073},
  year={2021}
}

@article{imageresponse1,
  title={Photochat: A human-human dialogue dataset with photo sharing behavior for joint image-text modeling},
  author={Zang, Xiaoxue and Liu, Lijuan and Wang, Maria and Song, Yang and Zhang, Hao and Chen, Jindong},
  journal={arXiv preprint arXiv:2108.01453},
  year={2021}
}

@article{imageresponse2,
  title={Multimodal dialogue response generation},
  author={Sun, Qingfeng and Wang, Yujing and Xu, Can and Zheng, Kai and Yang, Yaming and Hu, Huang and Xu, Fei and Zhang, Jessica and Geng, Xiubo and Jiang, Daxin},
  journal={arXiv preprint arXiv:2110.08515},
  year={2021}
}

@inproceedings{videogroundedqa1,
  title={Multimodal Transformer Networks for End-to-End Video-Grounded Dialogue Systems},
  author={Le, Hung and Sahoo, Doyen and Chen, Nancy and Hoi, Steven},
  booktitle={Proceedings of the 57th Annual Meeting of the Association for Computational Linguistics},
  pages={5612--5623},
  year={2019}
}

@inproceedings{videogroundedqa2,
  title={Audio visual scene-aware dialog},
  author={Alamri, Huda and Cartillier, Vincent and Das, Abhishek and Wang, Jue and Cherian, Anoop and Essa, Irfan and Batra, Dhruv and Marks, Tim K and Hori, Chiori and Anderson, Peter and others},
  booktitle={Proceedings of the IEEE/CVF Conference on Computer Vision and Pattern Recognition},
  pages={7558--7567},
  year={2019}
}

@article{videogroundedqa3,
  title={VGNMN: Video-grounded neural module network to video-grounded language tasks},
  author={Le, Hung and Chen, Nancy F and Hoi, Steven CH},
  journal={arXiv preprint arXiv:2104.07921},
  year={2021}
}

@inproceedings{videogroundedchitchat1,
  title={CHAMPAGNE: Learning Real-world Conversation from Large-Scale Web Videos},
  author={Han, Seungju and Hessel, Jack and Dziri, Nouha and Choi, Yejin and Yu, Youngjae},
  booktitle={Proceedings of the IEEE/CVF International Conference on Computer Vision},
  pages={15498--15509},
  year={2023}
}

@inproceedings{videogroundedchitchat2,
  title={TikTalk: A Video-Based Dialogue Dataset for Multi-Modal Chitchat in Real World},
  author={Lin, Hongpeng and Ruan, Ludan and Xia, Wenke and Liu, Peiyu and Wen, Jingyuan and Xu, Yixin and Hu, Di and Song, Ruihua and Zhao, Wayne Xin and Jin, Qin and others},
  booktitle={Proceedings of the 31st ACM International Conference on Multimedia},
  pages={1303--1313},
  year={2023}
}

@inproceedings{videogroundedchitchat3,
  title={Event-content-oriented dialogue generation in short video},
  author={Cheng, Fenghua and Li, Xue and Huang, Zi and Wang, Jinxiang and Wang, Sen},
  booktitle={Proceedings of the 2024 Conference of the North American Chapter of the Association for Computational Linguistics: Human Language Technologies (Volume 1: Long Papers)},
  pages={4114--4124},
  year={2024}
}

@inproceedings{videogroundedrealtime1,
  title={Livebot: Generating live video comments based on visual and textual contexts},
  author={Ma, Shuming and Cui, Lei and Dai, Damai and Wei, Furu and Sun, Xu},
  booktitle={Proceedings of the AAAI Conference on Artificial Intelligence},
  volume={33},
  number={01},
  pages={6810--6817},
  year={2019}
}

@inproceedings{videogroundedrealtime2,
  title={Videoic: A video interactive comments dataset and multimodal multitask learning for comments generation},
  author={Wang, Weiying and Chen, Jieting and Jin, Qin},
  booktitle={Proceedings of the 28th ACM International Conference on Multimedia},
  pages={2599--2607},
  year={2020}
}

@article{llmhallucination,
  title={A survey on hallucination in large language models: Principles, taxonomy, challenges, and open questions},
  author={Huang, Lei and Yu, Weijiang and Ma, Weitao and Zhong, Weihong and Feng, Zhangyin and Wang, Haotian and Chen, Qianglong and Peng, Weihua and Feng, Xiaocheng and Qin, Bing and others},
  journal={ACM Transactions on Information Systems},
  volume={43},
  number={2},
  pages={1--55},
  year={2025},
  publisher={ACM New York, NY}
}

@inproceedings{knowledgevqa1,
  title={Ok-vqa: A visual question answering benchmark requiring external knowledge},
  author={Marino, Kenneth and Rastegari, Mohammad and Farhadi, Ali and Mottaghi, Roozbeh},
  booktitle={Proceedings of the IEEE/cvf conference on computer vision and pattern recognition},
  pages={3195--3204},
  year={2019}
}

@inproceedings{knowledgevqa2,
  title={A-okvqa: A benchmark for visual question answering using world knowledge},
  author={Schwenk, Dustin and Khandelwal, Apoorv and Clark, Christopher and Marino, Kenneth and Mottaghi, Roozbeh},
  booktitle={European conference on computer vision},
  pages={146--162},
  year={2022},
  organization={Springer}
}

@inproceedings{multihopvqa1,
  title={Webqa: Multihop and multimodal qa},
  author={Chang, Yingshan and Narang, Mridu and Suzuki, Hisami and Cao, Guihong and Gao, Jianfeng and Bisk, Yonatan},
  booktitle={Proceedings of the IEEE/CVF conference on computer vision and pattern recognition},
  pages={16495--16504},
  year={2022}
}

@article{visionreasoning1,
  title={Selective Vision is the Challenge for Visual Reasoning: A Benchmark for Visual Argument Understanding},
  author={Chung, Jiwan and Lee, Sungjae and Kim, Minseo and Han, Seungju and Yousefpour, Ashkan and Hessel, Jack and Yu, Youngjae},
  journal={arXiv preprint arXiv:2406.18925},
  year={2024}
}

@inproceedings{visionreasoning2,
  title={Clevr: A diagnostic dataset for compositional language and elementary visual reasoning},
  author={Johnson, Justin and Hariharan, Bharath and Van Der Maaten, Laurens and Fei-Fei, Li and Lawrence Zitnick, C and Girshick, Ross},
  booktitle={Proceedings of the IEEE conference on computer vision and pattern recognition},
  pages={2901--2910},
  year={2017}
}

@article{rag,
  title={Retrieval-augmented generation for knowledge-intensive nlp tasks},
  author={Lewis, Patrick and Perez, Ethan and Piktus, Aleksandra and Petroni, Fabio and Karpukhin, Vladimir and Goyal, Naman and others},
  journal={Advances in neural information processing systems},
  volume={33},
  pages={9459--9474},
  year={2020}
}

@inproceedings{groundingdino,
  title={Grounding dino: Marrying dino with grounded pre-training for open-set object detection},
  author={Liu, Shilong and Zeng, Zhaoyang and Ren, Tianhe and Li, Feng and Zhang, Hao and Yang, Jie and others},
  booktitle={European Conference on Computer Vision},
  pages={38--55},
  year={2024},
  organization={Springer}
}

@article{Qwen2VL,
  title={Qwen2-VL: Enhancing Vision-Language Model's Perception of the World at Any Resolution},
  author={Wang, Peng and Bai, Shuai and Tan, Sinan and Wang, Shijie and Fan, Zhihao and Bai, Jinze and others},
  journal={arXiv preprint arXiv:2409.12191},
  year={2024}
}

@inproceedings{bleu,
  title={Bleu: a method for automatic evaluation of machine translation},
  author={Papineni, Kishore and Roukos, Salim and Ward, Todd and Zhu, Wei-Jing},
  booktitle={Proceedings of the 40th annual meeting of the Association for Computational Linguistics},
  pages={311--318},
  year={2002}
}

@inproceedings{meteor,
  title={METEOR: An automatic metric for MT evaluation with improved correlation with human judgments},
  author={Banerjee, Satanjeev and Lavie, Alon},
  booktitle={Proceedings of the acl workshop on intrinsic and extrinsic evaluation measures for machine translation and/or summarization},
  pages={65--72},
  year={2005}
}

@inproceedings{rouge,
  title={Rouge: A package for automatic evaluation of summaries},
  author={Lin, Chin-Yew},
  booktitle={Text summarization branches out},
  pages={74--81},
  year={2004}
}

@inproceedings{cider,
  title={Cider: Consensus-based image description evaluation},
  author={Vedantam, Ramakrishna and Lawrence Zitnick, C and Parikh, Devi},
  booktitle={Proceedings of the IEEE conference on computer vision and pattern recognition},
  pages={4566--4575},
  year={2015}
}

@inproceedings{dist,
  title={A Diversity-Promoting Objective Function for Neural Conversation Models},
  author={Li, Jiwei and Galley, Michel and Brockett, Chris and Gao, Jianfeng and Dolan, William B},
  booktitle={Proceedings of the 2016 Conference of the North American Chapter of the Association for Computational Linguistics: Human Language Technologies},
  pages={110--119},
  year={2016}
}

@inproceedings{bertscore,
  title={BERTScore: Evaluating Text Generation with BERT},
  author={Zhang, Tianyi and Kishore, Varsha and Wu, Felix and Weinberger, Kilian Q and Artzi, Yoav},
  booktitle={International Conference on Learning Representations},
  year={2019}
}

@article{ragapplication,
  title={Retrieval-augmented generation for large language models: A survey},
  author={Gao, Yunfan and Xiong, Yun and Gao, Xinyu and Jia, Kangxiang and Pan, Jinliu and Bi, Yuxi and Dai, Yi and Sun, Jiawei and Wang, Haofen and Wang, Haofen},
  journal={arXiv preprint arXiv:2312.10997},
  volume={2},
  year={2023}
}

@online{GeoGuessr,
  author = {GeoGUessr},
  title = {GeoGuessr - Let's explore the world},
  year = 2025,
  url = {https://www.geoguessr.com/},
  urldate = {2025-04-10}
}

@article{streetclip,
  title={Learning generalized zero-shot learners for open-domain image geolocalization},
  author={Haas, Lukas and Alberti, Silas and Skreta, Michal},
  journal={arXiv preprint arXiv:2302.00275},
  year={2023}
}

@article{geoclip,
  title={Geoclip: Clip-inspired alignment between locations and images for effective worldwide geo-localization},
  author={Vivanco Cepeda, Vicente and Nayak, Gaurav Kumar and Shah, Mubarak},
  journal={Advances in Neural Information Processing Systems},
  volume={36},
  pages={8690--8701},
  year={2023}
}

@article{deepseekvl2,
  title={Deepseek-vl2: Mixture-of-experts vision-language models for advanced multimodal understanding},
  author={Wu, Zhiyu and Chen, Xiaokang and Pan, Zizheng and Liu, Xingchao and Liu, Wen and Dai, Damai and others},
  journal={arXiv preprint arXiv:2412.10302},
  year={2024}
}

@inproceedings{llava1.5,
  title={Improved baselines with visual instruction tuning},
  author={Liu, Haotian and Li, Chunyuan and Li, Yuheng and Lee, Yong Jae},
  booktitle={Proceedings of the IEEE/CVF conference on computer vision and pattern recognition},
  pages={26296--26306},
  year={2024}
}


\end{document}